\documentclass[lettersize,journal]{IEEEtran}
\usepackage[utf8]{inputenc}

\usepackage{graphics}
\usepackage{graphicx}
\usepackage{color}
\usepackage{xcolor}
\usepackage{hyperref}
\hypersetup{colorlinks,breaklinks,citecolor=green!50!black,urlcolor=blue!50!black,linkcolor=red!50!black}
\usepackage{lscape}
\usepackage{booktabs}
\usepackage{xspace}
\usepackage{bbding}
\usepackage{pifont}
\usepackage{wasysym}
\usepackage{amssymb}
\usepackage[noadjust]{cite}
\newcommand{\nx}{\textcolor{red!70!black}{\ding{53}}}
\newcommand{\yx}{\textcolor{green!50!black}{\ding{51}}}

\newcommand{\etal}{{\em et al.}\xspace}
\usepackage{tikz}
\newcommand*\emptycirc[1][0.8ex]{\tikz\draw (0,0) circle (#1);} 
\newcommand*\halfcirc[1][0.8ex]{%
  \begin{tikzpicture}
  \draw[fill] (0,0)-- (90:#1) arc (90:270:#1) -- cycle ;
  \draw (0,0) circle (#1);
  \end{tikzpicture}}
\newcommand*\fullcirc[1][.8ex]{\tikz\fill (0,0) circle (#1);} \usepackage{xspace}
\newcommand{\BfPara}[1]{{\noindent \bf#1.\xspace}}

\title{Robust Natural Language Processing: Recent Advances, Challenges, and Future Directions}

\author{\IEEEauthorblockN{Marwan Omar}
\IEEEauthorblockA{University of Central Florida\\
marwan@knights.ucf.edu}
\and
\IEEEauthorblockN{Soohyeon Choi}
\IEEEauthorblockA{University of Central Florida \\
soohyeon.choi@knights.ucf.edu}
\and
\IEEEauthorblockN{DaeHun Nyang}
\IEEEauthorblockA{Ewha Womans University\\
nyang@ewha.ac.kr}
\and
\IEEEauthorblockN{David Mohaisen}
\IEEEauthorblockA{University of Central Florida\\
mohaisen@ucf.edu}
}

\author{Marwan Omar, Soohyeon Choi, DaeHun Nyang, and David Mohaisen 
\thanks{M. Omar, S. Choi, and D. Mohaisen are with the Department of Computer Science at the University of Central Florida, Orlando, Florida 32816, USA. D. Nyang is with Ewha Womans University, Seoul, Republic of Korea. D. Mohaisen is the corresponding author (e-mail: mohaisen@ucf.edu; phone: +1-407-823-1294).}}

\begin{document}

\maketitle

\begin{abstract}
     Recent natural language processing (NLP) techniques have accomplished high performance on benchmark datasets, primarily due to the significant improvement in the performance of deep learning. The advances in the research community have led to great enhancements in state-of-the-art production systems for NLP tasks, such as virtual assistants, speech recognition, and sentiment analysis. However, such NLP systems still often fail when tested with adversarial attacks. The initial lack of robustness exposed troubling gaps in current models' language understanding capabilities, creating problems when NLP systems are deployed in real life. In this paper, we present a structured overview of NLP robustness research by summarizing the literature in a systemic way across various dimensions. We then take a deep-dive into the various dimensions of robustness, across techniques, metrics, embeddings, and benchmarks. Finally, we argue that robustness should be multi-dimensional, provide insights into current research, identify gaps in the literature to suggest directions worth pursuing to address these gaps.

\end{abstract}
\section{Introduction}\label{sec:intro}

Over the last decade, machines talking and interacting with  humans in a human-like manner have become a reality. This reality is  seen in many human-facing and emerging applications, including  smart assistants, intelligent search engines, customer support, etc. Despite the significant differences among those applications in their associated contexts, they technically have one thing in common: they utilize an engine that employs advances in natural language processing (NLP) techniques, employing breakthroughs in deep machine learning (ML) and artificial intelligence (AI). As ML/AI continue to revolutionize our lives, leveraging and harnessing their power to understand natural languages, process and analyze them, and draw meaning from such analyses are the main promise that NLP applications aspire to deliver~\cite{mannarswamy2018evolving}. Technically, NLP today is a subarea of AI that allows machines read, understand, and obtain meanings from vast  language artifacts.

NLP?s key benefits lie at the heart of teaching computers how to analyze large amounts of textual data. Although it may seem like a new technology, with the emergence of recent successful applications, NLP's roots go back to the early 1950s when NLP was first used for machine translation (MT)~\cite{nadkarni2011natural}. With the technology wave we are experiencing leading to innovation and disruptive applications, the amount of text data being generated everyday grows exponentially. This, in turn, created the need for powerful technologies, such as NLP, for efficiently processing voluminous amounts of data. NLP is being widely adopted by many industries to provide meaningful interpretation for data and help solving numerous challenges~\cite{liu2020sentiment}. Moreover, NLP  applications can be seen in our everyday life, e.g.,  Google Translate, Google Assistance, Amazon Alexa,  Microsoft?s Cortana. In the financial industry, NLP is used in Prudential?s chat bot, Bank of America's Erica, among others. In the enterprise, NLP systems are  widely adopted for the detection of spam, intrusions, malware, etc. 

Machine learning techniques in general, and NLP techniques by association, are prone to attacks. Those attacks allow an  adversary to target those techniques, as part of the aforementioned applications, to violate the application objectives and guarantees~\cite{abusnaina2019adversarial}. For example, in smart speakers applications, it has been shown that an adversary use minimally-modified inputs to trigger wrong, and sometimes malicious, device activation through voice squatting~\cite{zhang2018understanding,blue2018hello,chen2017you,abuhamad2020sensor}. Similarly, an adversary might attack an NLP model that handles spam detection and fool it to make false predictions leading to spam passing through mail filters~\cite{abusnaina2019examining}. Malware authors might attack an NLP-based model to fool an intrusion detection system and missclassify malware as benign software~\cite{alasmary2020soteria}.  Adversaries might even be more tempted to attack NLP models making decisions on loans in the finance industry with the incentive to fool a loan application system to incorrectly qualify a customer for a loan or vise-versa~\cite{buber2017detecting}. 

The research community has produced various studies demonstrating that NLP models are vulnerable to adversarial (machine learning) attacks, as NLP models are susceptible to making incorrect predictions on adversarial examples \cite{morris2020textattack}. This, in turn, has led to a growing body of research on investigating and understanding the robustness of NLP techniques against adversarial attacks. Broadly speaking, such efforts in the literature are either focused on developing new attacks or better training models to make models resistant to such attacks (i.e., defenses)~\cite{goodfellow2014explaining}.  To sum up the research efforts dedicated understanding robustness in the literature, there are several research surveys that have addressed specific aspects of NLP robustness, e.g.,  data augmentation \cite{feng2021survey},  search methods \cite{yoo2020searching}, pretrained models \cite{li2021pretrained}, and adversarial attacks \cite{zhang2020adversarial}. However, the literature lacks research studies that provide a systematic overview of the state-of-the-art in this space across a range of variables; applications, technique, metrics, benchmark datasets, threat models, tasks, embedding techniques, learning techniques, goals, defense mechanisms, and performance. 

Motivated by the lack of a pipeline-oriented view of the literature in the domain of NLP robustness, we sample and provide a systematic overview of the body of work done thus far in this space addressing this problem, and the novel aspect is a categorization that goes beyond what has been done in the literature. The main objective of this effort is to provide a road-map to the existing work, particularly over the past few years, and the research gap that deserves further attention through investigation. We note that most of the work on NLP robustness in the past three years tackled the issue from one angle and provides partial solutions rather than a unified and comprehensive framework on how to fix weaknesses \cite{ribeiro2020beyond,zeng2021certified,ye2020safer,dong2020towards,kumar2020certifying,huang2019achieving,rychalska2019models,goel2021robustness,cheng2019evaluating,shi2020robustness,li2020textshield}.  Through this work, we wish to motivate the research community to develop a comprehensive frameworks to evaluate NLP robustness, in a pipeline  (i.e., as envisioned to be deployed in a real application, tackling various use model aspects and building blocks). In particular, such a framework should enable analysis and probing to disclose NLP-models' strengths and weaknesses and provide recommendations on how to address weaknesses. Moreover, we envision that any proposed solution should provide us the ability to visualize, analyze, and extensively test NLP models for robustness by utilizing state-of-the art tools, against a range of settings.

\BfPara{Contribution} The main goal of this work is a fresh and deep look into the recent work on NLP robustness. To this end, this work makes the following contributions.  (1) We introduce an enriched taxonomy that covers a range of dimensions of significant importance, driven from a pipeline of a broad range of NLP application.  (2)  We provide a categorization of various recent studies addressing NLP robustness, falling under the range of studied  variables; e.g., models, embedding techniques, metrics, and techniques, among others. 
(3)  We provide a contrast between the different approaches and their strengths and weaknesses. 
(4)  We provide a road-map of the gaps left by the existing literature and call for actions. 

Overall, this work offers researchers the ability to seek robustness from numerous aspects, e.g., choice of learning technique/model, embedding technique, datasets, defense mechanisms, and robustness metrics. 

\BfPara{Organization} The rest of the paper is organized as follows. In~\autoref{sec:related}, we review the related work in the context of NLP robustness (surveys). In \autoref{sec:overview}, we present an overview of a generic pipeline to guide our review. In \autoref{sec:techniques}, we provide a detailed review of the robustness techniques explored in the literature, which is the central theme of this work. In \autoref{sec:metrics}, we discuss the various metrics used for assessing robustness. In \autoref{sec:defense}, we highlight defence mechanisms for NLP. In \autoref{sec:embedding}, we discuss the impact of embedding on robustness, while the impact of dataset is covered in \autoref{sec:dataset}. We conclude in \autoref{sec:conclusion}.

\section{Related work}\label{sec:related}
This paper is a survey in nature, and there has been several surveys addressing robustness of NLP techniques, as mentioned earlier. However, those surveys are narrow in scope, and address only a narrow aspect of the robustness spectrum. For the completeness of treatment of the subject, we address those related surveys in the following. 

Feng \etal ~\cite{feng2021survey} conducted an extensive survey on data augmentation for NLP robustness. They studied various data augmentation techniques, including rule-based and model-based techniques as strategies to robustify NLP models to adversarial attacks. However, their work is limited in the sense that it only addresses a narrow aspect of robustness. Data augmentation is only part of several defense mechanisms to robustify NLP systems. For instance, there is robust data training and certifiable data training techniques, all of which are defense mechanisms to achieve robustness. 

Yoo \etal ~\cite{yoo2020searching} presented review of search algorithms for generating adversarial examples as a means to achieve robustness. However, their work is limited in scope in that it only focuses on adversarial examples as a means to seek robustness. There are many other variables in the robustness landscape such as embedding technique, robustness metrics, and robustness techniques which were not covered under their survey. Lin \etal~\cite{lin2021survey} presented a research survey on transformers which are pre-trained NLP models. The study is centered around robustness from a model stand point where learning models such as BERT and RoBERTa can contribute to robustness. However, their work falls short in addressing other aspects, e.g., defenses, attacks, and techniques. 

In \cite{zhang2020adversarial}, Zhang \etal conducted a survey on adversarial attacks as a means to evaluate robustness of NLP systems to adversarial perturbations. However, their work is limited due to the fact that robustness is multi-dimensional and adversarial attacks are only one dimension of robustness. Robustness entails numerous other elements such as defenses, metrics, and embedding technique. To the best of our knowledge, there is no comprehensive work in the literature that puts together advances on understanding the robustness considering a pipeline that accounts for the important steps in implementing an NLP system, which is our take in this work.

\begin{figure}[t]
    \centering
    \includegraphics[width=0.49\textwidth]{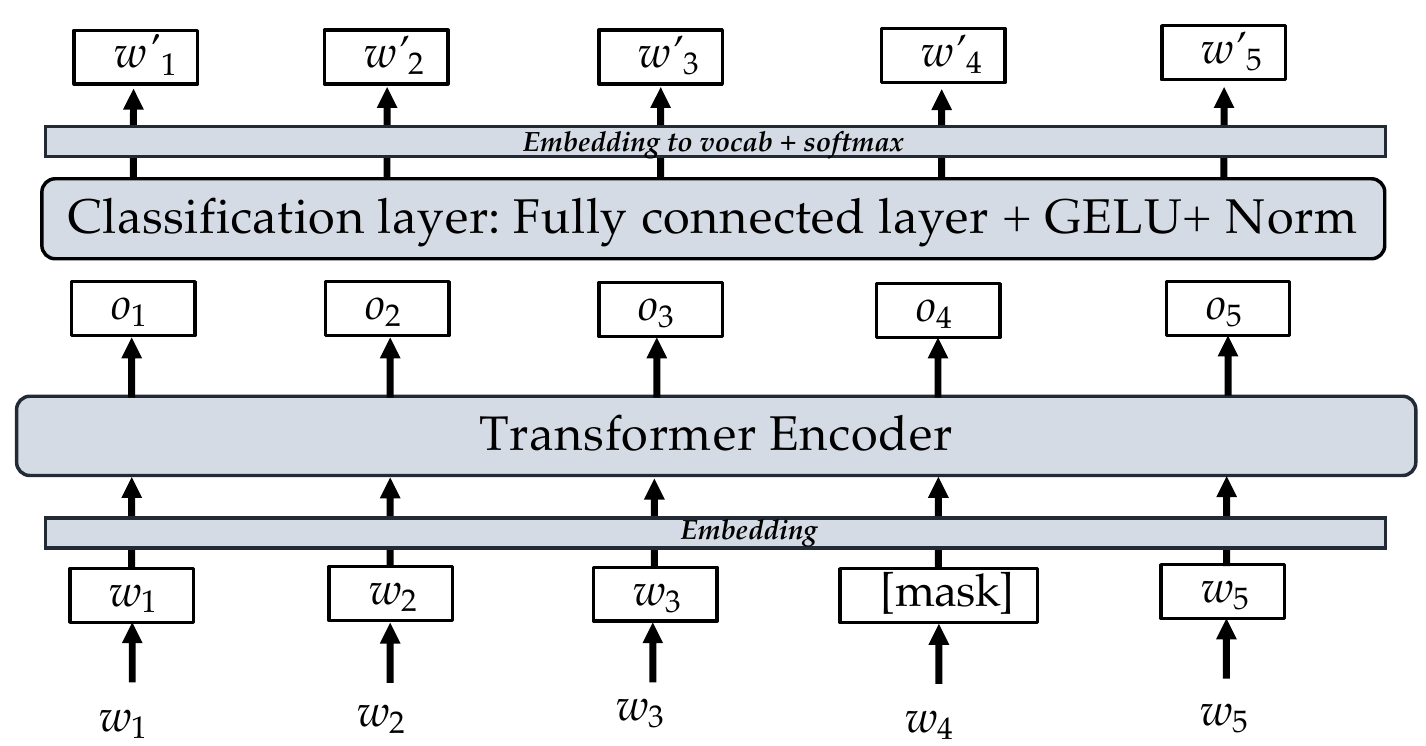}
    \caption{A typical NLP pipeline, using the BERT embedding technique.} 
    \label{fig:BERT Diagram}
\end{figure}

 \begin{figure*}[t]
    \centering
    \includegraphics[width=0.99\textwidth]{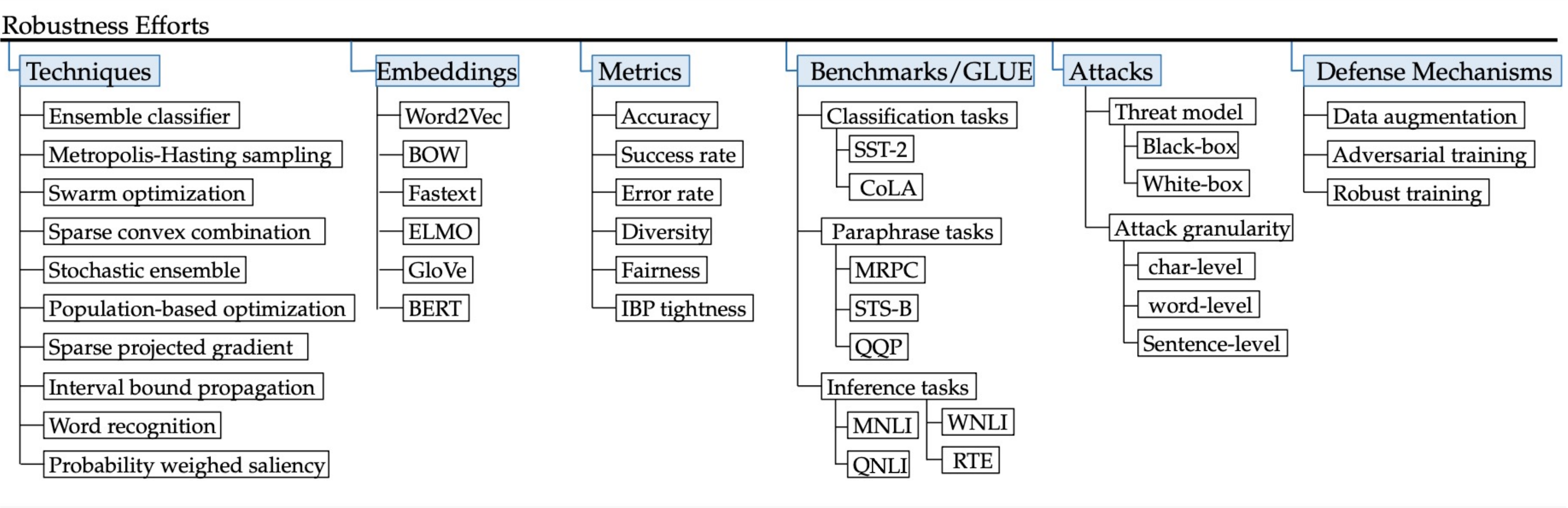}\vspace{-4mm}
    \caption{A high-level overview of the various research efforts in the domain of robustness analysis across various elements of the NLP pipeline, including techniques, embedding, metrics, benchmarks, attack model, and defense mechanisms.}
    \label{fig:NLP Diagram}\vspace{-4mm}
\end{figure*}


\section{NLP: A generic overview}\label{sec:overview}
To simplify our presentation of the overview of the various advances made over the past decade in the area of NLP towards improving our understanding of robustness through attacks and defenses, we highlight a system flow with various elements that are typical in NLP systems. We use those elements to describe the different advances. In the subsequent description, we envision the application of a natural language model (used for natural language generation).

At a high-level, and is exemplified later in \autoref{fig:BERT Diagram}, the typical NLP pipelines consist of a preprocessing step that takes a raw language input and prepares it for consumption by an NLP model through the appropriate steps of mapping. Upon that initial step of preprocessing and mapping, an embedding step is invoked to transform the initial representation into an appropriate format that can be consumed by the model. The model then runs, initially in a training phase, to learn several parameters that could be used for the language generation.



\BfPara{NLP Robustness: Taxonomy} In \autoref{fig:NLP Diagram}, we demonstrate a brief taxonomy of the various efforts presented in the literature on NLP and associated robustness analysis across the associated pipeline, including techniques, embedding, evaluation metrics, evaluation benchmarks (datasets), attack space (threat model and granularity), and associated defense mechanisms.


Given the range of objectives that each paper tries to address, there is a need to systematically understand them by breaking them down into some normal form, based on the pipeline we described here. As such, in the following sections, we dive into efforts that have been dedicated on each among those elements of the pipeline. 

To make this survey more accessible, we compile the acronyms used in the rest of this paper in \autoref{tab:Acronyms}.

\begin{table}[htb]
    \centering
        \caption{List of acronyms in alphabetical order.}
    \label{tab:Acronyms}
    \begin{tabular}{l|llll}
    \toprule
    Term & Definition \\\midrule
    AGNEWS & AG's News Topic Classification Dataset\\
    AI & Artificial Intelligence \\
    ASCC & Adversarial Sparse Convex Combination \\
    BERT & Bidirectional Encoder Representations from Transformers\\
    BoW & Bag-of-Words\\
    CNN & Convolutional Neural Networks \\
    CoLA & Corpus of Linguistic Acceptability \\
    DNE & Dirichlet Neighborhood Ensemble\\
    ELMO & Embeddings from Language Model \\
    GELU & Gaussian Error Linear Unit \\
    GloVe & Global Vectors for Word Representation \\
    GLUE  & Global Language Understanding Evaluation\\
    IB & Information Bottleneck \\
    IBP & Interval Bound Propagation\\
    IMDB & Internet Movie Database\\
    LSTM & Long Short Term Memory\\
    MCMC & Markov Chain Monte Carlo\\
    MH & Metropolis-Hastings\\
    MHA & Metropolis-Hastings Sampling Algorithm\\
    MHS & Metropolis-Hastings Sampling\\
    ML & Machine Learning\\
    MNLI & Multi-Genre Natural Language Inference \\
    MRPC & Microsoft Research Paraphrase Corpus \\
    MT & Machine Translation\\
    MTL & Multi-Task Learning\\
    NER & Named Entity Recognition\\
    NLI & Natural Language Inference\\
    NLP & Natural Language Processing\\
    OOD & Out-Of-Distribution\\
    PI & Paraphrase Identification\\
    PSO & Particle Swarm Optimization Algorithm \\
    PWWS & Probability Weighted Word Saliency\\
    QA & Question-Answering\\
    QNLI & Question-Answering Natural Language Inference\\
    QQP & Quora Question Pairs\\
    RG & Robustness Gym\\
    RoBERTa & Robustly Optimized BERT Pretraining Approach \\
    RTE & Recognizing Textual Entailment \\
    SA & Sentiment Analysis\\
    SEA & Semantically Equivalent Adversaries\\
    Seq2Seq & Sequence-to-Sequence\\
    SQuAD & Stanford Question Answering Dataset \\
    SPGD & Sparse Projected Gradient Descent \\
    SST-2 & Stanford Sentiment Treebank -- version 2\\
    STS-B & Semantic Textual Similarity Benchmark\\
    WNLI & Winograd Natural Language Inference \\
    \bottomrule
    \end{tabular}\vspace{-4mm}
\end{table}

\section{Robustness Techniques and Tools}\label{sec:techniques}


\subsection{Background}
As we pointed out in \autoref{sec:intro}, NLP models are being increasingly deployed to handle many human-like tasks. With the prevalence of NLP systems and their various real-world applications, the need for building robust NLP models becomes paramount because the consequences of making false predictions can be detrimental and even life-threatening in some cases (e.g., medical imaging diagnosis systems). In reality, however, we have seen numerous examples of failed NLP models after deployment in the real-world due to the lack of robustness. Some recent work has shown that approximately two-thirds of real-world NLP systems (e.g., essay grading NLP, Microsoft twitter NLP) fail after deployment due to the lack of robustness \cite{ribeiro2020beyond}. As a case in point, Amazon built an NLP-based recruiting tool that was deemed as a failure (and was eventually scraped) because the NLP-powered model demonstrated bias against female applicants~\cite{fabbri2021summeval}.

To address these issues, the security and NLP communities alike invested a significant amount of effort in developing techniques and tools for testing and analyzing the robustness of language processing techniques, embedding techniques that would provide better robustness, metrics to assess the performance, benchmarks to evaluate this robustness, attacks to challenge the robustness, and the associated defense mechanisms. In the following, we sample and review efforts in each of those directions, using the reference taxonomy in \autoref{fig:NLP Diagram}.

\subsection{Robustness Analysis and Testing Tools}
As a result of the failure of various NLP systems, the research community has conducted numerous studies on NLP robustness, calling for extensive testing of NLP models before deployment. For example, Rebeiro \etal~\cite{ribeiro2020beyond} introduced CheckList, a task-agnostic method for studying NLP models. CheckList incorporates a matrix of general linguistic capabilities and test types that allow for comprehensive test iteration. The proposed approach works on both commercial as well as research NLP models, and reveals model weaknesses even after models' internal testing, although stops short of providing solutions for the identified weaknesses. 

Similarly, Goel \etal \cite{goel2021robustness} identified challenges in evaluating NLP systems. As a result, they introduced a solution called Robustness Gym (RG), which is a simple yet extensible evaluation toolkit. By realizing a common mean for evaluation, RG enables NLP practitioners to compare various results from various frameworks and to develop new methods using a built-in sets of abstractions. Moreover, RG offers a unified NLP-model evaluation framework that allows thorough and extensive analysis and test of NLP models. While promising, the robustness framework does not seem to offer insights into the full understanding of the behavior of NLP models nor disclosing where the systems are actually failing or performing well. It would seem both desired and perhaps intuitive to extend the tool to localize the model failure and provide reasons behind model degradation.

In \cite{rychalska2019models}, Rychalska \etal introduced WildNLP, a framework for testing model stability in-situ. In WildNLP, text corruptions, such as keyboard errors or misspelling occur, are addressed. To this end, the authors compare the robustness of models in four popular NLP tasks: QA, NLI, NER, and SA. The authors do so by testing the performance of these tasks on aspects introduced in the framework, and find that the high performance of models does not guarantee sufficient robustness, although recent embedding techniques can help improve that. In order for us to improve the models robustness, we need to incorporate several factors rather than just simply using the adversarial attacks as a metric; e.g., the underlying model properties, test data, appropriate metrics, etc. 

Datasets and their role in highlighting the performance of various algorithms, as well as unveiling their robustness, have been also examined. In \cite{hendrycks2020pretrained}, Hendrycks \etal systematically examined and measured the out-of-distribution (OOD) generalization for seven NLP datasets. En route, they construct a robustness benchmark that employs realistic distribution shifts and measure the generalization of various models, including Bag-of-Words (BoW) models, CNNs, and LSTMs. Moreover, they show that the performance of pretrained transformers? decline is substantially smaller. The authors also examined the factors that affect the robustness, and found that larger models are not necessarily more robust than smaller models, while more diverse pretraining data could improve the robustness. The authors also use the generalization benchmark to train a model on the SST-2 \cite{socher2013recursive} dataset and evaluate on the IMDB \cite{jang2020bi} (both of which are popular benchmark datasets for the sentiment analysis task). They use BoW, CNN, and LSTM-based models to predict a movie review?s binary sentiment, and report the accuracy.

\subsection{Techniques for Robustifying NLP Models}

A number of techniques have been proposed in the literature for robustifying NLP models, including ensemble classifiers with randomized smoothing, stochastic ensembles, interval bound propagation, word recognition techniques, etc. 
In the following, we review some of the most widely used robustness techniques and where they are used. A summary of some of those works is shown in  \autoref{tab:robustness techniques}. Additionally, the reader is referred to~\autoref{fig:NLP Diagram} for additional context.

\begin{table*}[htb]
    \centering
        \caption{A comparison between various works from the literature, across techniques, whether a benchmark is utilized or not, the threat model (white-box vs black-box), and study's goal (attack, defense or robustness).}
    \label{tab:robustness techniques}\vspace{-2mm}
    \begin{tabular}{rllcccccc}
    \toprule
    Study &Year&Technique & Benchmark & White-Box & Black-Box & Attack & Defense & Robustness\\\midrule
 Carllini \etal \cite{carlini2017towards}	&2017&	Defensive Distillation	&	\yx	&	\yx	&	\nx	&	\nx	&  \nx& \yx	\\\hline
 Ebrahimi \etal \cite{ebrahimi2017hotflip}  &2017 & Character Substitution & \yx & \yx & \nx & \yx  & \yx &\yx \\\hline 

 Madry \etal \cite{madry2017towards}	&2017&	Projected Gradient Descent 	&	\yx	&	\yx	&\nx&\nx&\nx & \yx\\\hline
 Wong \etal \cite{wong2017dancin} & 2017 & Text Example Generation & \yx & \nx & \yx & \yx&  \nx& \nx\\\hline
 Zhao \etal \cite{zhao2017generating} & 2017 &  Stochastic Search Algorithm & \yx & \nx & \yx & \yx & \nx & \nx\\\midrule 

Alzantot \etal \cite{alzantot2018generating}	&2018&	Population-based Optimization 	&	\yx	&	\nx	&	\yx	&	\yx & \nx& \nx	\\\hline
Dong \etal \cite{dong2020towards}	&2018&	Sparse Convex Combination	&\nx		&	\nx	&	\yx	&	\nx	& \nx & \yx  	\\\hline

Geiger \etal\cite{geiger2018stress}	&2018&	Multiply-Quantified Sentences 	&	\nx	&	\yx	&\nx&\yx&\yx&\nx	\\\hline
Glochner \etal \cite{glockner2018breaking}	&2018&	Generating an NLI Test Set 	&	\yx	&	\yx	&	\yx	&\yx &\nx &\nx 	\\\hline

Gong \etal \cite{gong2018adversarial}	&2018&	Nearest-neighbour Search	&	\yx	&	\yx	&	\nx	&	\yx & \yx &\nx	 \\\midrule
Braham \etal \cite{barham2019interpretable}	&2019&	Sparse Projected Gradient Decent 	&\nx&	\yx	&	\nx	&\nx	&  \nx&\yx 	\\\hline
Clark \etal \cite{clark2019don}	&2019&	Learned-mixin Ensemble	&\nx		&	\yx	&	\nx	&\nx	&\yx& \nx	\\\hline

Eger \etal \cite{eger2019text}	&2019&	Character Substitution	&	\nx	&	\yx	&	\nx	&		\yx &\nx & \nx	\\\hline
He \etal \cite{he2019unlearn}	&2019&	Debiasing Data	&	\yx	&	\yx	&	\yx	&\yx&\yx&	\nx\\\hline
Jia \etal \cite{jia2019certified}	&2019&	Interval Bound Propagation 	&	\yx	&	\yx	&	\nx	&\nx	&\nx & \yx 	\\\hline
Jung \etal \cite{jung2019earlier}	&2019&	Biases in Summerization Analysis	&	\yx	&	\yx	&	\yx	&\nx&\nx&\yx\\\hline

Kaushik \etal \cite{kaushik2019learning}	&2019&	Spurious Correlation	&	\yx	&	\yx	&\nx&\yx&\yx&\nx	\\\hline
Pruthi \etal \cite{pruthi2019combating}	&2019&	Word Recognition	&	\yx	&	\yx	&\nx		&\nx	& \yx & \nx	\\\hline
Ren \etal \cite{ren2019generating}	&2019&	Probability Weighted Word Saliency	&	\yx	&	\yx	&	\nx	&		\yx & \nx&\nx 	\\\hline
Yaghoobzadeh \etal \cite{yaghoobzadeh2019increasing}	&2019&	Forgettable Examples for Robustness &\nx&\nx&	\yx&\nx& \nx &\yx\\\hline
Zang \etal \cite{zang2019word}	&2019&	Swarm Optimization Algorithm	&	\yx	&	\yx	&\nx		&		\yx & \nx&\nx 	\\\midrule
Cheng \etal \cite{cheng2020seq2sick} 	&2020&	Seq2Seq	&	\nx	&	\yx	&	\nx	&		\yx &\nx & \nx	\\\hline
Hendrycks \etal \cite{hendrycks2020pretrained}	&2020&	Out of Distribution (OOD)	&	\yx	&	\yx	&\nx&\nx& \nx & \yx 	\\\hline
Jin \etal \cite{jin2020bert}	&2020&	Text Generation	&	\yx	&	\nx	&	\yx	&		\yx & \nx& 	\nx\\\hline
Wang \etal \cite{wang2020infobert}	&2020&	Information Bottleneck (IB) Regulizer 	&\nx		&	\yx	&	\nx	&\nx&  \nx& \yx	\\\hline
Wang \etal \cite{wang2020identifying}	&2020&	Supervised Text Classification	&	\yx	&	\yx	&\nx&\yx&\yx&\nx	\\\hline
Wang \etal\cite {wang2020cat}	&2020&	Controlled Generation	&	\nx	&	\yx	&	\nx	&	\yx & \yx &	\nx\\\hline
Ye \etal\cite{ye2020safer}	&2020&	Stochastic Ensemble 	&	\yx	&	\nx	&	\yx	&\nx	& \yx & \nx	\\\hline

Zhang \etal \cite{zhang2020generating}	&2020&	Metropolis-Hastings Sampling 	&	\yx	&	\yx	&	\yx	&	\yx & \yx &	\nx\\\hline
Zhout \etal \cite{zhou2020defense}  	&2020&	Ensemble Classifier  	&\yx&	\yx	&	\nx	&\nx	& \yx & \nx	\\\midrule
Fabbri \etal \cite{fabbri2021summeval}	&2021&	Text Summerization	&	\yx	&	\yx	&\nx&	\yx & \yx &\nx\\\hline
Goel \etal \cite{goel2021robustness}	&2021&	Heuristics	&	\nx	&	\yx	&\nx&	\nx & \nx &\yx\\\hline

Schiller \etal \cite{schiller2021stance} &2021&Stance Detection Benchmark &  \nx & \yx	&\nx&	\yx & \yx &\yx\\\hline
Xu \etal \cite{xu2021grey} & 2021 & Text Generation & \yx	&	\yx	&\yx&	\yx & \yx &\yx\\\hline
Zeng \etal  \cite{zeng2021certified}	&2021&	RanMask	&	\yx	&	\nx	&	\yx	&	\nx	&  \nx& \yx 	\\\bottomrule
    \end{tabular}\vspace{-4mm}
\end{table*}

\subsubsection{Ensemble Classifiers  with Randomized Smoothing}
One of the obvious caveats of relying on a single classifier in NLP tasks is that a manipulation of the underlying input space (feature) fed into this classifier would have a significant impact on the output of the classifier. To cope with this issue, ensemble classifiers are proposed in the machine learning literature, where multiple classifiers (estimators) are built independently and aggregated to obtain the final result of the classifier. 
As such, and without losing generality, an ensemble classifier is a classifier whose decision depends on the combined outcome of decisions made by several individual classifiers, and is a method for achieving a degree of robustness in NLP models by reducing bias in the training data. 

By the same token, robustness with the ensemble classifier means that for any input $x$ and class label $y$, a smoothed classifier ($g$)  will return a prediction $g(x)$ which is most likely the correct prediction \cite{cohen2019certified}. As such, a model is said to be robust at $y$ if it can classify all inputs in the perturbation text correctly~\cite{jia2020building}. Randomized smoothing, on the other hand, is a method through which we can transform a classifier into a new smoothed classifier that is robust in a given setting.   Randomized smoothing can provably certify the robustness of NLP models against various adversarial attacks such as word-substitution attacks~\cite{kumar2020certifying}. For instance, Zhout \etal  proposed Dirichlet Neighborhood Ensemble (DNE), a randomized smoothing method for training a robust model in a way that mitigates substitution-based attacks~\cite{zhou2020defense}.  Essentially, DNE forms virtual sentences by sampling embedding vectors for each word in an input sentence from using a group of the word and its synonyms, and augments them with the training data. This sampling, in turn, ensures robustness against adversarial attacks without sacrificing the performance. 
While the randomized smoothing technique greatly enhanced classification accuracy against adversarial attacks, this technique only applies to one task at a time. In other words, if the technique was to be applied to another task, e.g., NLI, the robust training procedure would need to be restarted from scratch. This, in turn, will incur a significant amount of overhead \cite{jia2020building}.

\subsubsection{Stochastic Ensemble with Randomized Smoothing}
Stochastic ensemble refers to a classifier with some randomness and uncertainty in the underlying model. NLP models, in general, have a stochastic characteristic which (if understood correctly) enables us to effectively characterize the behavior of the NLP predictive models~\cite{zeng2021certified}.  In \cite{ye2020safer}, Ye \etal proposed a certified robustness method based on a new randomized smoothing technique that constructs a stochastic ensemble by applying random word substitutions on the input sentences. Moreover, their method leverages the statistical properties of the ensemble to provably certify robustness. 

This method is simple and generalizable in the sense that it does not depend on any structure and only requires black-box queries of the model outputs. As such, their method can be applied to any pre-trained model (e.g., BERT) and granularity (e.g., word-level, subword-level). Although {\em robust training} has been proven to enhance the overall robustness of the model against adversarial word-level perturbations, robust training on a different model requires re-executing the training steps from scratch which is one of the key limitations of certifiably robust training, and is an issue that this study fails to reveal. 
 
\subsubsection{Interval Bound Propagation}\label{sec:ibp}
The Interval Bound Propagation (IBP) is a technique used to build certifiably robust machine learning classifiers. IBP essentially uses the interval arithmetic to define a loss to minimize an upper bound on the maximal difference between any pair of logits when the input is perturbed within any norm-bounded ball \cite{gowal2018effectiveness}. IBP has been applied widely and successfully in the vision domain to obtain robustness guarantees \cite{ryou2020fast, kenmogne2018interval, gowal2018effectiveness,sridhar2021mitigating, lyu2021towards}. 

The key strength of IBP for NLP  models is that it can be used to process discrete perturbations in addition to the continuous ones, which are used in the computer vision domain~\cite{jia2020building}. It is our belief the  robustness cannot be understood in isolation of models obtained from different datasets, irrespective of the metric. As such, one obvious limitation of this line of work is that it did not use a broad benchmark; e.g., GLUE, which includes six benchmark datasets. Those datasets could have been possibly used to further evaluate the classification accuracy and demonstrate their robustness technique. 

\subsubsection{Word Recognition}
 Word recognition refers to an elementary process of language, whereby written and verbal forms of words are converted into linguistic tasks and representations.   Word recognition as a technique has been used in various studies~\cite{pruthi2019combating,dong2021towards,moradi2021evaluating, cui2021approach, gui2021textflint, bernier2020hardeval}. For example, Pruthi \etal proposed using a word recognition model in front of the classifier to combat adversarial spelling mistakes against a BERT model used for sentiment analysis. They show that a single adversarially-chosen character attack has lowered the accuracy from 90.3\%  to 45.8\%, while their defense brings the accuracy back up to 75\%. 
 
 This approach can be easily used to robustify NLP models against adversarial spelling mistakes. Moreover, the same approach can be  used for recognizing words corrupted by random keyboard mistakes, thus defending NLP  models against word perturbation attacks. What is unique about this approach is that, unlike many other studies which only study adversarial attacks on NLP models, it demonstrates vulnerable state-of-the-art NLP model and also proposes a robustness technique that contributes to protection against the same attacks. A study that combines both attack and defense strategies is certainly plausible. On the other hand, a limitation of this approach is that it is unclear if it transfers and generalizes to other network architectures across various linguistic tasks.

\subsection{Breaching Security by Improving Attacks}
Adversarial attacks have been used in the literature to evaluate robustness of NLP systems to real-world attacks. As depicted in \autoref{fig:NLP Diagram}, most research works tackle the adversarial attack issue from either an attack-granularity angle (character-level, word-level, and sentence-level attacks) or from a threat model (white-box and black-box) angle.

Numerous research studies have extensively studied the role of adversarial attacks in developing robust NLP models \cite{wang2020cat,ebrahimi2017hotflip,alzantot2018generating,cheng2020seq2sick}. For example, Cheng \etal~\cite{cheng2020seq2sick} study crafting AEs for seq2seq models whose inputs are discrete text strings. To address the challenges caused by the discrete input space, the authors propose a projected gradient method combined with group lasso and gradient regularization in the white-box threat model. To handle the large output space, they design a new loss functions that works for deriving both non-overlapping and targeted keyword attacks. The authors achieve an average of 85\% success rate with their adversarial attacks on NLP models, however, they do not indicate what attributed to the success of the attack: whether it is the poorly designed seq2seq model or is it the dataset. Also, authors stop short in offering any recommendations on how to increase the robustness of seq2seq models against adversarial attacks. 

From an attack granularity perspective, various studies have been carried out using either a character-level, word-level, or sentence-level attack \cite{wang2020cat,zang2019word,zhang2020generating,wong2017dancin,zhao2017generating,gao2018black,eger2019text}.  For example, Eger \etal~\cite{eger2019text} investigate the impact of visual adversarial attacks (modification to text which can be detected by visualizing) on  character-, word-, and sentence-level tasks. They show that both neural and non-neural models, in contrast to humans, are vulnerable to such attacks, leading to a performance decrease of up to 82\%. 
In the following subsection, we explore, in details, some of the attack methods/techniques used in the NLP robustness literature:



\subsubsection{Adversarial Sparse Convex Combination (ASCC)}
Sparse convex combination refers to the method of representing the target output as a sparse convex combination of the input text. Based on this definition, for any input $x$ and class label $y$, a trained NLP classification model maps each input $x$ to its class label $y$. Given a clean (unperturbed) input $x$ a targeted sparse adversarial attack aims for finding a perturbation so that the perturbed input $x'$ is incorrectly classified to a target class \cite{dong2020towards}. This attack method has been used in numerous research studies \cite{dong2020towards,blum2018x,szeghy1adversarial,dong2021towards,jia2017adversarial,yuan2021sparsegan,tsiligkaridis2020understanding}. 

For example, in \cite{dong2020towards}, Dong \etal introduced an Adversarial Sparse Convex Combination (ASCC) method to model the word substitution attack space and leverage a regularization term to enforce perturbation towards an actual substitution. In doing so, they align their modeling better with the discrete textual space. Based on the ASCC method, they also generate worst-case perturbations and incorporate  adversarial training for robustness. Their experiments show that ASCC-defense outperforms the current state-of-the-art techniques in terms of robustness on two prevailing NLP tasks, SA and NLI, and address several attacks across several architectures.

The strength of this attack method is that it can be used to generate adversarial examples to robustify models and eventually improve models' prediction accuracy. On the other hand, when robust accuracy on adversarial examples goes up, this causes the clean accuracy (unperturbed standard test data) to go down, a trade-off that should be considered when utilizing this attack method \cite{jia2020building}.  

\subsubsection{Population-based Optimization for Adversarial Attacks}

A population-based optimization algorithm is a type of genetic algorithm that aims to find perturbations which can change a model?s prediction/classification~\cite{birbil2004convergence,khormali2020generating}. This technique maintains a ``population'' of candidates' inputs and continuously perturbs and combines them~\cite{jia2020building}. On the other hand, a black-box attack is a type of adversarial attack where an adversary does not have access to the model's internal structure nor parameters. This attack technique has been used in numerous research works \cite{alzantot2018generating, zang2019word, maheshwary2021generating, li2021turning, morris2020textattack,hui2020foolchecker,larijani2020adversarial,gupta2020determining,jasser2021resilience,suzuki2019adversarial}. 

Alzantot \etal~\cite{alzantot2018generating} proposed a population-based optimization algorithm to generate semantically and syntactically similar AEs that can fool SA and textual entailment models with high accuracy. Moreover, they demonstrate that more than 90\% of the successful SA AEs are classified to their original label by 20 human annotators, and that the examples are perceptibly quite similar. 

While this attack technique concretely examines the vulnerability of NLP models to adversarial examples, it falls short of proving whether the uncovered attack remains effective under various model architectures (transferability). For instance, it is unknown if the attack success rate would remain the same under the LSTM  and Word-CNN models. 
 
\subsubsection{Sparse Projected Gradient Descent}
The projected gradient descent method is a type of greedy algorithm which has been applied broadly to machine learning models~\cite{madry2017towards}. In this method,  each element in the input text is considered for substitution and the best perturbations are selected from all possible perturbations and rerun until no more perturbations are possible \cite{yoo2020searching}.  This attack method has been utilized in several research works \cite{yoo2020searching, wang2019towards, barham2019interpretable} with various promising results. For example, Barham \etal~\cite{barham2019interpretable} introduced a sparse projected gradient descent (SPGD) method for crafting interpretable AEs for text applications. SPGD imposes a regularization constraint on input perturbations by projecting them onto the directions to nearby word embeddings with the highest similarity. 

The strength of this attack method is that the regularization constraint ensures that perturbations move each word embedding in an interpretable direction (i.e., towards another nearby word embedding) while ensuring a high prediction accuracy on different model architectures. A limitation of this attack method is that it is implemented using only one dataset, which is the IMDB. Ideally, robust NLP models should be evaluated across various linguistic tasks on multiple datasets. Research has shown that models should be tested using the GLUE (Global Language Understanding Evaluation) benchmark which includes nine datasets, six of which are for testing classification accuracy~\cite{wang2019towards}.  

\subsubsection{Probability Weighted Word Saliency (PWWS)}
PWWS is a greedy search method for generating adversarial examples~\cite{erkan2004lexrank}. In this method, the goal is to rank words based on some importance function. In a descending importance order, each word is replaced with a candidate word until we successfully perturb all words \cite{yoo2020searching}. This technique has been used in multiple studies addressing machine learning robustness to adversarial attacks, including the studies in~\cite{ma2020adversarial,pasunuru2018multi,ding2019saliency,jeon2018discovering,ren2019generating}. The way such a technique is used in those studies is almost identical. For example, Ren \etal~\cite{ren2019generating} addressed the problem of AEs on text classification by generating AEs that maintain lexical and grammatical correctness, as well as semantic similarity. Based on the synonyms substitution strategy, they introduce a word replacement order determined by both the word saliency and the classification probability, and propose a greedy PWWS algorithm for text AEs. Their experiments on three datasets using convolutional and LSTM based models show that their approach reduces the classification accuracy  and keeps a very low word substitution rate. 
The strength of this technique is that it exposes the vulnerabilities of NLP models to adversarial examples via word replacement using an efficient greedy algorithm. One limitation of this technique, however, is that it does not consider the error analysis aspect of robustness. In other words, the technique does not indicate which models were correct on the original words/data but incorrect on the perturbed words. 

\subsubsection{Swarm Optimization Algorithm for Adversarial Attacks}

The Particle Swarm Optimization Algorithm is a search algorithm used to generate adversarial examples~\cite{zang2019word}. In this method, each member of the population is perturbed by creating all potential candidate obtained by replacing each input and then sampling one input example, at each iteration.  Using this algorithm, we are able to find the best perturbed input among all members of the population \cite{yoo2020searching}. This attack technique has been used in multiple studies addressing machine learning robustness to adversarial attacks in general, including the studies in~\cite{zang2019word,zang2020learning, roth2021token,yoo2020searching,morris2020textattack}. The way such a technique is used in those studies is almost identical; e.g., Zang~\etal \cite{zang2019word} propose an attack model that incorporates a word substitution method~\cite{qi2019openhownet} and particle swarm optimization-based search algorithm for that purpose with a significant success. 

The strength of this method is its generalization to different model architectures, such as BiLSTM and BERT, using benchmark datasets. Moreover, this attack method achieves higher attack success rates and crafts more high-quality AEs in comparison to various baseline methods. 

A key limitation of this attack method, however, is that it does not take into account run-time of the PSO algorithm, as the run-time is a critical factor for search algorithms in any real-world deployment. Per~\cite{yoo2020searching}, a key factor to consider for those algorithms' complexity is the length of the input text, as well as the choice of the search algorithm. For instance, if the input texts are short (e.g., a few sentences), a beam search is an appropriate choice, since it can achieve a high success rate without incurring the overhead.  In such tasks, AEs must be generated quickly, and a more efficient algorithm may be preferred, even with a lower success rate.

\subsubsection{Metropolis-Hastings Sampling for Adversarial Attacks} The Metropolis Hastings (MH) Sampling is a Markov Chain Monte Carlo (MCMC) algorithm for generating a sequence of random samples from a probability distribution where direct sampling is hard \cite{zhang2020generating}. MH works by conducting a random walk according to a Markov chain whose stationary distribution is $\pi$ (the eventual distribution from which the chain will sample). On each step of the MC, a new state is proposed and either accepted or rejected according to a dynamically calculated probability value, called the acceptance criteria \cite{jia2017adversarial}. That is, in the long term, the data points from the MC will look similar to the data points from $\pi$ \cite{zhou2020defense}, \cite{zhang2020generating}. 

This attack technique has been used in multiple studies addressing machine learning robustness to adversarial attacks, including \cite{alshemali2020generalization,wang2020hamiltonian,liu2020collaborative,zhang2020generating}. The way such a technique is used in those studies is almost identical. For example, Zhang \etal propose Metropolis-Hastings Sampling (MHS) to generate fluent adversarial examples for attacking NLP models. The authors perform Metropolis-Hastings sampling which is designed with the guidance of gradients. The strength of this method is that it outperforms the baseline models on attacking capability. However, this method was not tested under various model architectures as well as linguistic tasks. Ideally, an attack technique should scale up to re-use across numerous tasks (e.g., sentiment analysis and NLI) and with various model architectures (e.g., RoBERTa, LSTM, Attention-based, etc.).

\subsection{Insights and Open Directions}

It is evident that the attack methods utilized in the literature for evaluating the performance and robustness of NLP models are diverse. This diversity of attack methods is dictated by the variety of limitations that some of those attacks have, precluding their use in broad set of applications while advocating others for their strengths and ease of implementation. In general, NLP models have to be evaluated to determine their robustness to adversarial attacks, where different attacks may prove more potent than others. 

Overall, our exploration of the attack methods space calls work in various directions to fill various gaps. (1) While there is a significant initial work on the utilization and implementation of various attack techniques in the broad NLP community, the attacks developed so far are limited in many ways, and there is a need for developing techniques that are transferable and can be generalized to various NLP model architectures. (2) Generally, and with a few exceptions, the majority of the work in the literature considers specific task definitions for which the robustness is analyzed and understood with respect to the proposed attacks. For instance, many of such efforts have focused on only a specific task, such as  sentiment analysis, question answering, etc. and left unaddressed the challenge of defending NLP models against a generic adversary optimizing in the input language for  multiple tasks~\cite{geiger2018stress,glockner2018breaking,garg2020bae,kaushik2019learning,jung2019earlier}. Moreover, most research works focus on how to develop certain types of concrete adversarial examples in a constrained adversarial setting. This creates an open challenge, that is hopefully within reach, for the research community where attack techniques should consider an advanced adversary and take into account how a determined and unconstrained adversary might circumvent robustness measures put for NLP tasks. (3) State-of-the-art NLP systems are demonstrably vulnerable to adversarial attacks which, in turn, causes inaccuracies in their prediction capabilities. Although researchers have provably shown the vulnerable state of NLP models, they fall short, however, in identifying the gaps in models' capabilities and how to improve such models. It remains an open challenge for the  NLP research community to conduct extensive experiments to deep-dive into the internal structure of NLP models to distinguish models which fare better or worse than others under adversarial attacks. For instance, it would be interesting to know if a BERT model fares better or worse than an LSTM (Long Short Term Memory) model.
(4) Most of the research progress on NPL robustness techniques concerning the development of new attacks takes into consideration  worst-case scenarios/examples without indicating which type of worst-case scenario to focus on. While that is understanding from a security standpoint, the comprehensive best-case and average-case analysis would be worthwhile, although unclear how tractable such an analysis would be. 
\section{Robustness Metrics}\label{sec:metrics}
The robustness of NLP models is a quality that has to be measured with well-defined and relevant metrics in order to gain an understanding of  the level of NLP model's resistance against adversarial attacks. In general, metrics serve a dual purpose in machine learning systems design: measuring their performance in training and testing. Robustness metrics are similar, in the sense that they are used for measuring and tracking the performance of the machine learning models under adversarial settings. Based on the surveyed literature, we found that different research works utilize different metrics for measuring robustness. A summary of some of those works is shown in Table II, with the context highlighted in~\autoref{fig:NLP Diagram}. In the following, we review some of those metrics.  

\begin{table*}[ht]
    \centering
        \caption{A listing of robustness metrics.  \fullcirc~means robustness is satisfied, \halfcirc~is partially satisfied, and \emptycirc~for not satisfied. The metrics we use are the accuracy rate, success rate, error rate, IBP accuracy, Perturbation size, fairness, sensitivity, and diversity.}
    \label{tab:metrics}\vspace{-2mm}
    \scalebox{0.9}{
    \begin{tabular}{rcccccc|ccccc}
    \toprule
    Paper	&Year &	\rotatebox[origin=c]{0}{Robustness}	&
\rotatebox[origin=c]{0}{Accuracy}&	
\rotatebox[origin=c]{0}{Success} 	&	
\rotatebox[origin=c]{0}{Error} 	&	
\rotatebox[origin=c]{0}{IBP}	&	
\rotatebox[origin=c]{0}{Perturbation}	&	
\rotatebox[origin=c]{0}{Fairness}	&
\rotatebox[origin=c]{0}{Void}	&
\rotatebox[origin=c]{0}{Sensitivity} &
\rotatebox[origin=c]{0}{Diversity} \\\midrule

Carllini \etal \cite{carlini2017towards} & 2017	&\fullcirc	&	\yx	&	\yx	&	\nx	&\nx		&\nx		&	\nx	&		&	\nx	&\nx		\\\hline

Ebrahimi \etal \cite{ebrahimi2017hotflip}& 2017	&	\halfcirc	&	\yx	&\nx&\nx&\nx&	\yx	&\nx&  &\nx&\nx\\\hline
Hosseini \etal \cite{hosseini2017deceiving} & 2017	&\halfcirc	&	\yx	&	\nx	&	\yx	&\nx		&\yx		&	\nx	&		&	\nx	&\nx		\\\hline
Liang \etal \cite{liang2017deep} & 2017	&\emptycirc	&	\yx	&	\nx	&	\yx	&\nx		&\yx		&	\nx	&		&	\nx	&\nx		\\\hline
Wong \etal \cite{wong2017dancin} & 2017	&\emptycirc	&	\yx	&	\nx	&	\yx	&\nx		&\nx		&	\nx	&		&	\nx	&\nx		\\\midrule
Geiger \etal\cite{geiger2018stress} &2018 &\halfcirc	&	\yx	&	\yx	&	\yx	&\nx		&\nx		&	\nx	&		&	\nx	&\nx		\\\hline
Glochner \etal \cite{glockner2018breaking} &2018 &\emptycirc	&	\yx	&	\nx	&	\nx	&\nx		&\yx		&	\nx	&		&	\yx	&\yx		\\\hline
Iyyer \etal \cite{iyyer2018adversarial}&2018 &\fullcirc	&	\yx	&	\nx	&	\yx	&\nx		&\nx		&	\nx	&		&	\nx	&\nx		\\\hline
Naik \etal \cite{naik2018stress}	& 2018&	\halfcirc	&	\yx	&	\nx	&	\nx	&	\nx	&	\nx	&	\nx	&		&	\nx	&	\nx	\\\hline
Ribeiro \etal \cite{ribeiro2018semantically} & 2018&	\fullcirc	&	\yx	&	\nx	&	\nx	&	\nx	&	\nx	&	\nx	&		&	\nx	&	\nx	\\\midrule

Braham \etal \cite{barham2019interpretable}& 2019	&	\halfcirc	&	\yx	&\nx&\nx&\nx&	\yx	&\nx&    &\nx&\nx\\\hline
Chang \etal \cite{chang2019bias}&2019 &\halfcirc	&\nx&	\nx	&\nx&\nx&\nx&\yx&   &\nx&\nx\\\hline

Cheng  \etal \cite{cheng2019evaluating}& 2019	&\fullcirc	&\nx&	\yx	&\nx&\nx&\nx&\nx&   &\nx&\nx\\\hline
Eger \etal \cite{eger2019text}	& 2019&	\halfcirc	&	\yx	&\nx&\nx&\nx&\nx&\nx&   &\nx&\nx\\\hline

Haung \etal \cite{huang2019achieving}& 2019	&	\fullcirc	&	\yx	&\nx&\nx&	\yx	&\nx&\nx&		&\nx&\nx\\\hline
Huang \etal \cite{huang2019reducing}&2019 &\emptycirc	&\nx&	\nx	&\nx&\nx&\nx&\yx& &\nx&\nx\\\hline
Ilyas \etal \cite{ilyas2019adversarial}& 2019	&\fullcirc	&	\yx	&	\nx	&	\nx	&\nx		&\nx		&	\nx	&		&\nx		&	\nx	\\\hline

Jia \etal \cite{jia2019certified}	& 2019&	\fullcirc	&	\nx	&	\nx	&	\nx	&	\yx	&	\yx	&\nx		&		&\nx		&\nx		\\\hline
Pruthi \etal \cite{pruthi2019combating}& 2019	&	\fullcirc	&\nx		&	\nx	&	\yx	&	\nx	&	\nx	&	\nx	&		&	\yx	& \nx		\\\hline
Ren \etal \cite{ren2019generating}	& 2019&		\emptycirc	&	\yx	&\nx&\nx&\nx&\nx&\nx&   &\nx&\nx\\\hline
Rychalska \etal \cite{rychalska2019models}& 2019	&	\halfcirc	&	\yx	&\nx&\nx&\nx&\nx&\nx&   &\nx&\nx\\\hline
Zhu \etal \cite{zhu2019freelb}	& 2019&	\halfcirc	&	\yx	&\nx		&	\nx	&	\nx	&	\nx	&	\nx	&		&\nx		&	\yx	\\\hline

Zang \etal \cite{zang2019word}	& 2019&	\emptycirc	&\nx&\yx	&\nx&\nx&\nx&\nx&   &\nx&\nx\\\midrule

Cheng \etal \cite{cheng2020seq2sick} & 2020	&	\emptycirc	&\nx&	\yx	&\nx&\nx&	\yx	&\nx&   &\nx&\nx\\\hline
Dong \etal \cite{dong2020towards}& 2020	&	\fullcirc	&	\yx	&\nx&\nx&\nx&\nx&\nx&  &\nx&\nx\\\hline
Dong \etal \cite{dong2020api}	& 2020&	\halfcirc	&	\yx	&\nx&\nx&\nx&\nx&\nx&   &\nx&\nx\\\hline
Jin \etal \cite{jin2020bert}& 2020	&	\emptycirc	&	\nx	&	\nx	&	\nx	&	\nx	&	\yx	&	\nx	&   &\nx&\nx\\\hline
Li \etal \cite{li2020textshield}& 2020	&	\fullcirc	&	\yx	&\nx&\nx&\nx&\nx&\nx&   &\nx&\nx\\ \hline
Moriss \etal \cite{morris2020textattack}& 2020	&		\emptycirc	&	\yx	&\nx&\nx&\nx&	\yx	&\nx&  &\nx&\nx\\\hline
Shi \etal \cite{shi2020robustness}	& 2020&	\fullcirc	&\nx		&\nx		&\nx		&	\yx	&	\nx	&	\nx	&		&\nx		&	\nx	\\\hline
Sharma \etal \cite{sharma2020data}	& 2020&	\fullcirc	&	\yx	&	\nx	&	\nx	&	\nx	&	\nx	&	\yx	&	&	\nx	&	\nx	\\\hline
Wang \etal \cite{wang2020infobert}	& 2020&	\fullcirc	&\nx&\nx&\nx&\nx&\nx&\nx&  	&\nx&\nx\\\hline
Wang \etal\cite {wang2020cat}	& 2020&	\halfcirc	&	\yx	&	\nx	&	\nx	&	\nx	&	\nx	&	\nx	&		&	\nx	&	\yx	\\\hline
Ye \etal\cite{ye2020safer}	& 2020&	\fullcirc	&	\yx	&	\nx	&	\nx	&	\nx	&	\nx	&	\nx	&		&	\nx	&\nx		\\\hline
Yoo \etal \cite{yoo2020searching}	& 2020&	\halfcirc	&\nx		&	\yx	&\nx		&\nx		&	\yx	&	\nx	&	&	\nx	&\nx		\\\hline
Zhang \etal \cite{zhang2020generating}	& 2020&	\halfcirc	&	\yx	&	\nx	&\nx		&	\nx	&\nx		&	\nx	& 	&\nx		&	\nx	\\\hline
Zhout \etal \cite{zhou2020defense}  & 2020	&	\halfcirc	&	\yx	&	\nx	&	\nx	&	\nx	&	\yx	&\nx		&		&	\nx	&\nx		\\\midrule

Czarnowska \etal \cite{czarnowska2021quantifying}	& 2021&	\emptycirc	&	\yx	&\nx&\nx&\nx&\nx&\yx&   &\nx&\nx\\ \hline
Feng \etal \cite{feng2021survey}	& 2021&	\fullcirc	&	\yx	&\nx&\nx&\nx&\nx&\nx&   &\yx&\yx\\ \hline
Goel \etal \cite{goel2021robustness}	& 2021&	\halfcirc	&	\yx	&\nx&\nx&\nx&\nx&\nx&   &\nx&\nx\\ \hline
Xu \etal \cite{xu2021grey}	& 2021&	\fullcirc	&	\yx	&\nx&\nx&\nx&\nx&\nx&   &\yx&\yx\\ \hline
Zeng \etal  \cite{zeng2021certified}& 2021	&	\halfcirc	&	\nx	&	\yx	&	\nx	&	\nx	&	\yx	&	\nx	&		&\nx		&	\nx	\\\bottomrule

    \end{tabular}}\vspace{-4mm}
\end{table*}

\subsection{Attack Success Rate}
Attack success is one of the simplest and most widely utilized metrics for evaluating the robustness of NLP models. The attack success rate refers to the number of attempts that are successfully normalized by the number of overall attempts of an attack (e.g., number of valid adversarial examples that both meet a predefined example condition on the size of perturbation and the adversary's objective; e.g., reducing the confidence of a classifier below a given threshold, or changing the classification label of the example). 

As a metric, the attack success rate has been utilized in numerous research studies to determine the effectiveness of adversarial attacks on NLP models \cite{deng2020analysis, morris2020textattack, yoo2020searching, gardner2020evaluating, alzantot2018generating}. For instance, Alzantot \etal \cite{alzantot2018generating} measured the effectiveness of their genetic algorithm-based adversarial attacks using the attack success rate as a metric which eventually indicates NLP model robustness to adversarial examples.  

While the attack success rate is simple and easy to interpret metric, its main disadvantage is that it ignores most, if not all, quality characteristics of the resulting adversarial examples that contributed to the success rate. For instance, as we will see later, not all adversarial examples are considered of the same quality, where some of them may be easily eliminated or detected using simple heuristics while others are more challenging to address using the same heuristics. 


\subsection{Error Rate}

Error rate (also known as the robustness error) refers to the number of  times where an NLP model incorrectly classifies an input text. The error rate is a metric which has been used in numerous research studies to determine the robustness of NLP models to adversarial attacks~\cite{sharma2017attacking}. In contrast to the attack success rate, the lower the error rate (misclassification rate), the more robust the NLP model is against adversarial attacks. 

The error rate has been used as metric in numerous research studies, including \cite{liang2017deep,hosseini2017deceiving, wong2017dancin, goodfellow2014explaining, iyyer2018adversarial}. For example, Goodfellow \etal~\cite{goodfellow2014explaining} found that several models, including neural network-based models, consistently misclassify AEs inputs formed by applying small but intentionally worst-case perturbations to the input examples from a dataset. In doing so, the perturbed input forces the model to output an incorrect answer with high confidence. According to the authors, adversarial examples are often misclassified by a variety of classifiers with different architectures.

This metric is simple and easy to calculate in order to evaluate the robustness of NLP models to adversarial attacks. However, the error rate alone should not be the only metric to evaluate the performance of machine learning models as it does not take into account the intrinsic and often clear differences between the examples contributing to the error rate. 

\subsection{IBP Bounds Tightness}
As highlighted in \textsection\ref{sec:ibp}, IBP is a technique used to accomplish robustness. Researchers studied the tightness of IBP's upper and lower bounds as a metric to determine and formally verify the degree of model robustness against adversarial attacks \cite{li2014high}. A model achieves a provably-guaranteed robustness against an attack if it cannot cross the boundary, no matter how adversaries create adversarial examples \cite{wang2019towards}. 

The IBP tightness metric has been utilized in several research works~\cite{huang2019reducing, jia2019certified, shi2020robustness, xu2020automatic}. For example, in \cite{shi2020robustness}, Shi \etal used the IBP tightness to study the robustness verification problem for transformers. In \cite{jia2019certified}, Jia \etal used the same metric to study certified robustness to word substitutions and considered an exponentially large family of label-preserving transformations where each word in the input text can be swapped with a similar one. 
The advantage of using the IBP tightness metric is that it can be used to evaluate verifiable robustness of NLP models to word substitution attacks. On the other hand, this metric alone should not be used as an indication of the true certified robustness. Ideally, other evaluation metrics should be used in conjunction with IBP tightness, such as normal accuracy and training accuracy.
\subsection{Classification Accuracy}
The classification accuracy is a simple extension of the accuracy metric, and refers to NLP model's ability to correctly classify input texts under different attack methods (e.g., white-box and black-box attacks, or word-level and character-level substitution attacks, among other settings) \cite{derczynski2016complementarity}. 

The classification accuracy has been utilized by numerous research works~\cite{zhang2020generating, carlini2017towards, ebrahimi2017hotflip, hosseini2017deceiving, geiger2018stress, glockner2018breaking, naik2018stress, iyyer2018adversarial, eger2019text}. For example, in \cite{zhang2020generating}, Zhang \etal used the classification accuracy metric to evaluate their proposed Metropolis-Hastings Sampling Algorithm (MHA) and demonstrated that MHA under classification accuracy outperforms the baseline model on attacking capability. 

Similar to other accuracy measures, this metric is easy to calculate and interpret, providing an ideal mean for easily comparing different algorithms. However, on the downside, this metric is agnostic to the quality of the individual examples contributing to the classification accuracy.



\subsection{Diversity}
Diversity implies that examples from training data of one class are as differentiable as possible from training data of another class to promote invariance in training data \cite{jia2019certified}. 

The diversity metric has been utilized by numerous research studies \cite{barham2019interpretable, cheng2019evaluating, clark2019don, derczynski2016complementarity, zhu2019freelb} to measure the classification accuracy of NLP models as part of the robustness to adversarial attacks. For example, in \cite{zhu2019freelb}, Zhu \etal propose a new adversarial training algorithm called FreeLB, which provides a higher invariance in the embedding space by perturbing input words to minimize the resulting adversarial risk on the input text. For validation, they apply their approach to transformer-based models in both NLU and reasoning tasks. Their experiments show that, when applied only to the fine tuning stage, their approach is able to improve the overall test score of BERT-based model from 78.3\% to 79.4\%, and RoBERTa-large model from 88.5\% to 88.8\%. The authors, however, stop short in explaining how to measure the run-time of their algorithm because the performance and accuracy of search algorithms sometimes become a trade-off issue. There are many search algorithms in the literature such as genetic algorithm, particle swarm optimization, greedy search, among other. Most of these algorithms have been thoroughly studied and could have been considered for this research as well \cite{yoo2020searching}.

One strength of the diversity metric is that it supports a qualitative notion across NLP domains making it suited for  evaluating model in ways that are not exhibited in any of the prior metrics.  Moreover, this metric allows us to understand how the NLP models  generalize in predicting validation target features. However, it is unclear how this metric can be used to measure diversity via precision and recall.
\subsection{Fairness}
Fairness in the context of machine learning, and NLP in particular, refers to the fair representation of data points for a particular language understanding task. This metric also aims to ensure that the NLP models do not make erroneous assumptions to produce prejudice results. As a case in point, Amazon?s job candidate NLP-based system was deemed to have prejudice against female applicants because the NLP model did not have a fair representation  for female applicants? resumes (i.e., it was not trained with enough female resumes). 

The fairness metric has been utilized by various studies \cite{chang2019bias, huang2019reducing, czarnowska2021quantifying, gautam2020sgg, sharma2020data} to evaluate the degree of NLP robustness and ensure that large models are fair.  Sharma \etal~\cite{sharma2020data} presented a simple data augmentation technique that selectively adds a subset of synthetic points in order to meet a fairness criterion without compromising the accuracy. Experiments are performed on three datasets where they have shown that their method outperforms prior methods to lessen bias while maintaining the accuracy. 

This metric can help reduce and control algorithmic biases and increase how fairly models perform in the real-world. The weakness of this metric is that it is coarse-grained, and cannot easily detect or even differentiate between the unintended biases that may exist in NLP models. Furthermore, it should not be the only measure to determine that whether algorithmic biases are completely removed from models or not.



\subsection{Insights and Open Directions}
It is clear that the metrics utilized in the literature for measuring the performance and robustness of NLP models are diverse. This diversity of metrics is necessitated by the variety of shortcomings that some of those metrics have, precluding their use in broad set of applications. Some others are advocated for their simplicity and ease of interpretation. 

In general, NLP models have to be evaluated to determine their robustness to adversarial attacks, where different metrics may prove more practical than others. The robustness can be evaluated by computing the upper and lower bounds of IBP, for instance. The upper bound  can be minimized using the back-propagation approach. The lower bound can be achieved using IBP to measure certified accuracy \cite{jia2017adversarial, rajpurkar2018know}. 

Our analysis of the surveyed work shows that NLP models are more robust to adversarial attacks (e.g., word substitution attacks) when trained with robust training (i.e., the IBP) as opposed to normally-trained models which fared poorly (classification accuracy of 36.0\%) under the same adversarial attacks \cite{jia2017adversarial}. An interesting area of research in this context is whether models trained with data augmentation would be more robust to attacks than robustly trained models. Another recommendation is to explore and experiment with clean versus robust accuracy metrics. We observed that robust training fared well (classification accuracy of up to 87\%) against adversarially perturbed words \cite{jia2017adversarial}, but we do not know if this robustness would cause any increase or drop in clean accuracy (accuracy on clean/unperturbed words). 

Overall, our exploration of the metrics space calls for work in various directions to fill various gaps. (1) While there is a significant initial work on the understanding of bias in the broad NLP community, the metrics developed so far are limited in many ways, and there is a need for developing techniques for assessing fairness and removing biases in data to evaluate how NLP models would perform when deployed to the real-wold. Namely, it would be interesting and worthwhile to extend the existing notions and metrics to techniques that address the existence of residual and unintended biases in datasets. (2) Diversity is understood in terms of the accuracy as a target metric, and it would be worthwhile to extend the diversity metrics to concretely evaluate models using the precision and recall as potential factors. (3) There seems to be a gap and need for techniques to allow the integration and use of IBP tightness metric in conjunction with other metrics such as training accuracy and normal accuracy. 

\section{Defense Mechanisms }\label{sec:defense}
Neural NLP systems must learn the fragile predictability of natural languages in order to address the generalization flaws of NLP systems \cite{jia2020building}. We have already seen how NLP models trained with standard data are vulnerable to adversarial attacks \cite{alzantot2018generating}. To address those vulnerabilities, a range of techniques have been studied in the literature, including robustness through data augmentation, adversarial training, and multi-task learning. We note that the literature has several surveys that address each individual technique, which we refer the reader to, although we highlight the high-level ideas of each of those techniques and exemplify the techniques by some sample works for the completeness in treating the subject.

\subsection{Data Augmentation}
In the computer vision domain, data augmentation and robust training, as defense mechanisms, have been proven to robustify neural models to adversarial perturbation \cite{zhang2019theoretically, madry2017towards}. Inspired by those developments in the vision field, NLP researchers have considered adversarial training, data augmentation, and robust training as defense mechanisms to robustify NLP models (as depicted in \autoref{fig:NLP Diagram}). For example, Zeng \etal~\cite{zeng2021certified} proposed a certifiably robust defense by randomly masking a certain words from the input to defend against both word substitution based attacks and character-level perturbations. The authors claim that they can certify the classifications of over 50\% texts to any perturbation of 5 words on AGNEWS dataset, and 2 words in the SST-2 dataset (dataset-dependent). The interested reader could find more details in the comprehensive survey on data augmentation techniques, their advantages, and disadvantages in~\cite{feng2021survey}. 

\subsection{Adversarial Training}
There has been also a significant body of work on the use of adversarial training in defending against attacks on NLP models and systems. It is noted that such attacks are not limited to this application domain, and are prevalent to most learning-based systems. 


Given the sufficiency of the prior survey work, and the pace of progress in this domain with respect to the NLP models and applications, the interested reader might refer to the survey of Chakraborty \etal for more related works~\cite{chakraborty2018adversarial}.

\subsection{Multi-Task Learning}
Multi-Task Learning (MTL) is a learning technique that enables researchers to share useful information or representations between and among related machine learning tasks. This technique is so popular that it has been adopted by researchers and practitioners across many domains including computer vision, speech recognition, and NLP tasks \cite{li2021empirical}. The benefit of sharing information from related tasks offers the ability to generalize deep learning models more efficiently on the original task.  The MTL technique has been utilized in numerous research studies including \cite{tu2020empirical, mccann2018natural, bingel2017identifying, subramanian2018learning} and the way this technique is used in those works is very similar. For example, in \cite{tu2020empirical}, Tu \etal proposed to use multi-task learning (MTL) to improve generalization as a form of robustness in NLP models. The authors experimented on NLI and paraphrase identification to show that MTL leads to significant performance gains. The authors demonstrated the importance of data augmentation and diversity for addressing spurious correlations challenges. The study was carried out on NLI and paraphrase identification (PI).

\subsection{Insights and Open Directions}
 We observe that data augmentation, adversarial training, multi-task learning, and robust training all have a positive impact on the classification accuracy of NLP models thereby contributing to robustness. We have also noticed from analyzing the literature that robust training outperforms both adversarial training as well as data augmentation when it comes to robustness to adversarial attacks \cite{jia2017adversarial}.
 
 Overall, our exploration of the defense mechanisms space calls for work in various directions to fill various gaps. (1) While there is a significant initial work on the utilization of adversarial training in the broad NLP community. The adversarial techniques available so far are limited in many ways, and there is a need for developing techniques that evaluate NLP models' robustness beyond deployment to the real-world. Namely, it would be interesting and worthwhile to quantify the impact of adversarial training on rapidly evolving language models as such models will be exposed to unsean data after deployment. (2) There seems to be a gap and the need for robust training techniques to test model robustness across various models because most of the research works in this context conduct robust training under a certain model architecture (e.g., BERT, Glove, etc.). To achieve robustness to adversarial attacks, an NLP model must be evaluated using more than one architecture on various datasets. For instance, in a sentiment analysis task, an NLP model should be evaluated using embeddings such as BoW, GloVe, Word2Vec, and RoBERTa on benchmark datasets from GLUE. (3) Although data augmentation has proven to increase model prediction accuracy, it has not been thoroughly examined to see its long-term impact on model performance. Because language models shift and drift after deployment to the real-work, it is paramount to develop techniques for reevaluating the impact of data augmentation on the long-run. Additionally, most of the research studies that leverage randomized smoothing with IBP, fail to consider the run time aspect, which is the overhead incurred during computation. We wish to motivate the NLP research community to consider the above gaps as future research directions. 
 
 \begin{table*}[t]
    \centering
        \caption{Representative literature work with various embedding techniques. \fullcirc~stands for robustness being satisfied, \halfcirc~for partially satisfied, and \emptycirc~for not satisfied.}
    \label{tab: Embedding models}\vspace{-2mm}
    \begin{tabular}{rccccccccc}
\toprule
{ Paper}	&	Year&
\rotatebox[origin=c]{0}{ Robustness}	&	
\rotatebox[origin=c]{0}{ BoW}&	
\rotatebox[origin=c]{0}{ Word2Vec} 	&	
\rotatebox[origin=c]{0}{ Glove} 	&	
\rotatebox[origin=c]{0}{ ELMO}	&	
\rotatebox[origin=c]{0}{ ROBERTA}	&	
\rotatebox[origin=c]{0}{ BERT}	&	
\rotatebox[origin=c]{0}{ LSTM}	\\\hline

Ebrahimi \etal \cite{ebrahimi2017hotflip}	&2017 &	\halfcirc	&\nx&\nx&\nx&\nx&\nx&\nx&	\yx	\\\hline
Hoseini \etal \cite{hosseini2017deceiving} &2017 &	\halfcirc	&\nx&\yx&\nx&\yx&\nx&\yx&	\nx	\\\hline
Jia \etal \cite{jia2017adversarial}&2017 &	\fullcirc	&\nx&\nx&\nx&\nx&\nx&\yx&	\yx	\\\hline
Wong \etal \cite{wong2017dancin}	& 2017&	\emptycirc	&\nx&\nx&\nx&\nx&\nx&\nx&	\yx	\\\hline
Zhao \etal \cite{zhao2017generating}	& 2017&	\emptycirc	&\nx&\nx&\nx&\nx&\nx&\nx&	\yx	\\\midrule

Alzantot \etal \cite{alzantot2018generating}	 &2018 &	\emptycirc	&\nx&\nx&	\yx	&\nx&\nx&	\yx	&\nx\\\hline
Gao \etal \cite{gao2018black} &2018 &	\halfcirc	&\yx&\nx&	\yx	&\nx&\nx&	\nx	&\nx\\\hline
Geiger \etal\cite{geiger2018stress}	& 2018&\emptycirc	&	\yx	&\nx&\nx&\nx&\nx&\nx&	\yx	\\\hline

Gong \etal \cite{gong2018adversarial}	& 2018&	\emptycirc	&\nx&\nx&	\yx	&\nx&\nx&\nx&\nx\\\hline

Naik \etal \cite{naik2018stress}	& 2018&	\halfcirc	&	\yx	&\nx&\nx&\nx&\nx&\nx&	\yx	\\\midrule

Braham \etal \cite{barham2019interpretable}	& 2019&	\halfcirc	&\nx&\nx&\nx&\nx&\nx&\nx&	\yx	\\\hline
Cheng \etal \cite{cheng2019evaluating}	& 2019&\fullcirc	&\nx&\nx&\nx&\nx&\nx&	\yx	&\nx\\\hline
Haung \etal \cite{huang2019achieving}&2019	&	\fullcirc	&\nx&\nx&\nx&\nx&\nx&	\yx	&\nx\\\hline

Jia \etal \cite{jia2019certified}	& 2019&	\fullcirc	&	\yx	&\nx&\nx&\nx&\nx&\nx&	\yx	\\\hline
Pruthi \etal \cite{pruthi2019combating}&2019	&	\fullcirc	&\nx&\nx&\nx&\nx&\nx&	\yx	&	\yx	\\\hline
Ren \etal \cite{ren2019generating}	& 2019&	\emptycirc	&\nx&\nx&	\yx	&\nx&\nx&\nx&	\yx	\\\hline
Rychalska \etal \cite{rychalska2019models}	 &2019 &	\halfcirc	&\nx&\nx&	\yx	&\nx&\nx&	\yx	&\nx\\\hline

Schmitt \etal \cite{schmitt2019sherliic}&2019	&	\emptycirc  &\nx&	\yx	&\nx&\nx&\nx&\nx&\nx\\\hline
Wallace \etal \cite{wallace2019universal}	 & 2019&	\fullcirc	&\nx&	\yx	&\nx&	\yx	&\nx&\nx&\nx\\\hline
Zang \etal \cite{zang2019word}	& 2019&	\emptycirc	&\nx&\nx&	\yx	&\nx&\nx&	\yx	&	\yx	\\\hline
Zhu \etal \cite{zhu2019freelb}	& 2019&	\halfcirc	&\nx&\nx&\nx&\nx&	\yx	&	\yx		&\nx	\\\midrule

Dong \etal \cite{dong2020towards} &2020			&	\fullcirc	&\nx&\nx&\nx&\nx&\nx&\nx&	\yx	\\\hline
Gardner \etal \cite{gardner2020evaluating}	& 2020&\emptycirc	&\nx&\nx&\nx&	\yx	&	\yx	&	\yx	&\nx\\\hline

Hendrycks \etal \cite{hendrycks2020pretrained}	 & 2020&	\halfcirc	&\nx&	\yx	&	\yx	&\nx&	\yx	&	\yx	&\nx\\\hline
Jin \etal \cite{jin2020bert} &2020	&	\emptycirc	&\nx&\nx&\nx&\nx&\nx&	\yx	&\nx\\\hline

Moriss \etal \cite{morris2020textattack}	&2020&	\emptycirc	&\nx&\nx&\nx&\nx&\nx&	\yx	&	\yx	\\\hline
Shi \etal \cite{shi2020robustness}	& 2020&	\fullcirc	&\nx&\nx&\nx&\nx&\nx&	\yx			
\\\hline
Wang \etal \cite{wang2020infobert}	&2020 &\fullcirc	&\nx&\nx&\nx&\nx&	\yx	&	\yx	&\nx\\\hline

Ye \etal\cite{ye2020safer}	&2020 &	\fullcirc 	&\nx&\nx&\nx&\nx&\nx&	\yx	&\nx\\\hline
Yoo \etal \cite{yoo2020searching}	&2020	&	\fullcirc	&\nx&\nx&	\yx	&\nx&\nx&	\yx	&	\yx	\\\hline
Zhang \etal \cite{zhang2020generating}	& 2020&	\halfcirc	&\nx&\nx&\nx&\nx&\nx&	\yx	&	\yx	\\\hline
Zhout \etal \cite{zhou2020defense} &2020 	&	\halfcirc	&	\yx	&\nx&\nx&\nx&\nx&\nx&	\yx	\\\midrule
Fabbri \etal \cite{fabbri2021summeval} &2021&	\fullcirc	&\nx&\nx&\nx&\nx&\nx&	\yx	&\nx\\\hline
Goel \etal \cite{goel2021robustness}	&2021&	\halfcirc	&\nx&\nx&\nx&\nx&\nx&	\yx	&\nx\\\hline
Schiller \etal \cite{schiller2021stance} &2021&	\fullcirc	&\nx&\nx&\nx&\nx&\nx&	\yx	&\nx\\\hline
Xu \etal \cite{xu2021grey}&2021&	\fullcirc	&\nx&\nx&\nx&\nx&\nx&	\yx	&\yx\\\hline
Zeng \etal  \cite{zeng2021certified}	 & 2021&	\halfcirc	&\nx&\nx&\nx&\nx&	\yx	&	\yx	&\nx\\\bottomrule
    \end{tabular}\vspace{-4mm}
\end{table*}

\section{Embedding Techniques}\label{sec:embedding}
Several embedding techniques have been utilized in the literature for representing text and natural language entities (in \autoref{fig:NLP Diagram}). The choice of those models is influenced by various factors, including their fitness to the studied applications, performance, and robustness. In the following, we review some of those embeddings and where they are used. A summary and contrast of some of those works are shown in \autoref{tab: Embedding models}.




\subsection{Representation Techniques}

\subsubsection{Bag of Words} The bag-of-words model, or BoW for short, is a technique used for extracting features from raw data for use in NLP models \cite{zhao2017fuzzy}. Moreover, this technique is considered as a popular text embedding technique widely used for text classification tasks such as sentiment analysis. In the BoW technique, the input text is represented as the bag of its words without regard to word order or grammar. 

The BoW embedding method has been implemented by numerous research studies \cite{geiger2018stress,naik2018stress, gao2018black, jia2020building,zhou2020defense} to achieve robustness for NLP tasks such a sentiment analysis and spam filtering. For example, in \cite{geiger2018stress}, Geiger \etal proposed a method for generating semantically challenging NLI data sets using the popular BoW embedding technique and showed that a range of NLI neural models (especially models based on the BoW embedding technique) invariably learn sub-optimal solutions and fail to encode crucial information. The authors concluded that certain NLP models are not fit for certain NLP tasks due to inherent weaknesses in their underlying architecture (hence the implementation of BoW). 

The strengths of the BoW technique are that: (1) it is easy to implement. (2) it offers flexibility and ability of customizing it to the specific task, text data type, and structure. However, on the downside, and because BoW ignores words ordering, it will lead to ignoring the context, which negatively impacts word meanings (semantics). The fact that word meaning and context are ignored by BoW, is a significant limitation of the model's linguistic capabilities.



\subsubsection{Word2Vec} Word2Vec is an model used to transform words into vectors. This is achieved by representing text into a numerical format that deep neural networks can understand \cite{jatnika2019word2vec}. This step is necessary to help NLP models understand text in the form of numbers. Word2vec (if provided with sufficient data and context) has the capability to make highly accurate predictions for tasks such as sentiment analysis (i.e., by classifying reviews with high certainty) and topic modeling. 

The Word2Vec technique has been utilized by numerous research studies including \cite{hosseini2017deceiving, wallace2019universal, schmitt2019sherliic, hendrycks2020pretrained} to reveal a model's weaknesses by showing a lower classification accuracy score. For instance, in \cite{wallace2019universal}, Wallace \etal introduced triggers as a new form of universal adversarial perturbation and used Word2Vec to evaluate the robustness of NLP models to adversarial attacks. The authors proposed a gradient guided search over tokens to finds trigger sequences that successfully lead to the target prediction. Their experiments utilized Word2Vec embedding technique and demonstrated a sharp decrease in accuracy of certain NLP models. 
The benefit of utilizing the Word2Vec representation technique is that it is simple and intuitive. Moreover, the way it is implemented enables the Word2Vec to easily learn how words are represented in classification tasks. Another advantage of this technique is that it does not require huge preprocessing, and hence less memory, because the model accepts data in an online manner. 
On the downside, this technique suffers from its inability to deal with out-of-vocabulary words, since Word2Vec model is unable to interpret unseen words. Moreover, this technique does not scale to other languages because it would require new embedding metrics.     


    

\subsubsection{GloVe} \underline{Glo}bal \underline{Ve}ctor for word representation is another representative learning technique which is a reformulation of the Word2vec optimization algorithm \cite{ibrahim2020enriching}. In this method, words are represented as vectors to create word co-occurrence matrices \cite{pennington2014glove}. The core concept in this method is to determine the frequency of word occurrence in the co-occurrence matrix. 

The GloVe learning technique has been utilized by numerous research studies ~\cite{hendrycks2020pretrained,alzantot2018generating,ren2019generating,gong2018adversarial, rychalska2019models} to study the impact of word learning representation on robustness. For instance, in \cite{rychalska2019models}, Rychalska \etal introduced WildNLP, a framework for studying model stability with text corruptions, e.g., keyboard errors and misspelling. In their pursuit, the authors use the GloVe embedding technique as a baseline to evaluate NLP robustness. One advantage of this learning technique is that it considers the frequency of co-occurrences (also refereed to as global statistics) crucial to building word embeddings. Thus word re-occurrences is directly tied to word vectors.  

Because GloVe is considered a count-based model, it potentially requires more computational power for more processing, a major draw-back of this technique.


\subsubsection{ELMO} Embedding from Language Models, or ELMO, is a word embedding technique for transforming a sequence of words into a sequence of vectors. In this approach, the input data is represented as character-level tokens and the output is word-level embeddings \cite{peng2019transfer}. We note that this learning technique differs from the previously mentioned learning techniques in that it computes vectors for an entire sentence instead of assigning a vector for each word embedding. This means that the same word can be assigned to different word vectors if the context is different. This is where the difference between ELMO and other traditional embedding techniques (e.g., GloVe, and Word2vec) come into play.  ELMO has been successfully utilized in numerous works, including \cite{hosseini2017deceiving, wallace2019universal, kaushik2019learning, gardner2020evaluating}, to accomplish high classification accuracy in the context of robustness for many NLP linguistic tasks including text classification and question-answering. 

One advantage of ELMO is that it can handle the out-of-distribution issue because of its ability to use character embeddings to represent word embeddings. However, on the downside, ELMO requires huge computation time to realize the word vectors \cite{gupta2020study}, a limitation we may overcome by pre-computing the vectors offline. 


\subsubsection{Fastext} This technique is an extension of the Word2Vec method: instead of transforming each word from the input text into a vector as output, this method represents each word as an n-gram of characters for the output. Fastext can be used to reveal a model's ``blind spots'' in the context of robustness and has been utilized for such purpose by numerous research studies \cite{ribeiro2018semantically,joulin2016fasttext, wang2019evaluating, santos2017sentiment, athiwaratkun2018probabilistic}. For instance, in \cite{ribeiro2018semantically}, Ribeiro \etal conducted a  study to find bugs in NLP models based on Fastext and presented a semantically equivalent adversaries  and semantic-preserving perturbations, defined as perturbations that produce changes in the model's predictions. The authors implemented their method using several NLP tasks; QA and SA. 

The benefit of using the Fastext learning technique is that it employs a simple and efficient baseline for sentence classification and uses N-gram features to reduce computation time and enhance efficiency. However, it might take a longer time to train a Fastext model because of the fact it uses N-gram features which could be greater than the number of words. This means that Fastext embedding is fit for only a certain linguistic tasks such as classification tasks.  



 \subsection{Insights and Open Directions}
It is evident that representation learning techniques utilized in the literature for evaluating NLP models are diverse. Impressive results on a certain learning technique should not imply that a given NLP model will perform equally well when deployed in the real-world. Overall, our exploration of those techniques calls for work in various directions to fill various gaps. (1) While there is a significant initial work on the utilization of various embedding techniques in the broad NLP community, the current research works fall short in using the same embedding technique/model (e.g., Fastext, BERT) across various linguistics tasks. (2) To harness the full power and potential of an embedding technique, there is a need for developing learning techniques, e.g., it would be interesting and worthwhile to  build an encoding layer (within the model's architecture) that can enforce resistance to perturbations. (3) There seems to be a gap in exploring the reuse of encoding across various linguistics tasks for robustness, which has the potential to improve classification accuracy. 


\section{Robustness via Benchmark Datasets}\label{sec:dataset}

NLP systems have performed remarkably on a wide spectrum of linguistic tasks due, in large, to the emergence of deep neural networks and unsupervised pre-training \cite{devlin2018bert}. Standard benchmark datasets have accomplished excellent results across many NLP tasks. For example, the SQuAD dataset has 95.4\% F-1 score which outperforms human accuracy \cite{yogatama2019learning}. 

As tempting as it may be to believe how well NLP models perform on standard datasets generally, often these models are actually solving the ``dataset problem'' rather than solving the underlying task with sufficient generalizations. For instance, in~\cite{yogatama2019learning}, a BERT model is trained on the SQuAD dataset and achieved an impressive 86.5\% F-1 score. We note, however, that these results are chiefly because the model is tested on a data that is created in the same way as the training data, following the same distribution and patterns. This, in turn, provides a false-sense of confidence in NLP model performance that may generalize for OOD samples. On the flip side, when the same model was tested on TriviaQA dataset (which is created in the same format as the SQuAD dataset), the F-1 score dropped to 35.6\%. 

It is important to understand the data utilized in validating and evaluating a task to reach conclusions on generalization. One such best practice to achieve this goal is to challenge the model with OOD samples. For instance, one approach to show this practice is by exposing the model to a training set, and not to samples from within the distribution of the testing set. One possible benefit of examining models with such settings is that surface cues can be identified to show limitations of models, as pointed out in a previous section, where gaps could be addressed by using techniques that can help with robustness and generalization, e.g., data augmentation techniques \cite{feng2021survey}.

Another approach to extend the generalization of models and improve the robustness to weak forms of attacks is to utilize widely-accepted and standard benchmark dataset: the fact that those datasets are standard imply that they went through rigorous evaluation for representation and soundness of collection. For instance, GLUE (General Language Understanding Evaluation), depicted in \autoref{fig:NLP Diagram}, is one of evaluation tools, which is a collection of nine benchmark datasets. GLUE is designed for analyzing and effectively evaluating NLP systems on three linguistic tasks, classification  (benchmark datasets: SST-2 and CoLA), paraphrasing (benchmark datasets: MRPC, STS-B, and QQP), and inference  (benchmark datasets: MNLI, WNLI, QNLI, and RTE) \cite{devlin2018bert}.

While benchmarks are an excellent way to improve the robustness and examine the generalization of models through well-vetted data, their main purpose is to mitigate bias and address spurious correlations, which we review in the following.




\subsection{Dataset Bias}
One of the benefits of utilizing rich benchmarks is addressing the explicit bias.  Dataset bias in NLP refers to a form of error in which certain elements (i.e., variables or attributes) of a dataset are more heavily represented than others \cite{hutchinson2020social}, this skewing the resulting NLP model recognizing such bias and affecting its operation against a more diverse data distribution. In other words, the biased dataset does not accurately reflect a model's true use case in the real-world, resulting in analytical errors and misleading classification accuracy levels. 

The issue of dataset bias in NLP has been studied extensively in numerous research works \cite{hutchinson2020social, he2019unlearn, clark2019don, shah2019predictive}. The way those research works examine data bias varies depending on the task and domain. For example, in \cite{he2019unlearn}, He \etal argued that NLI tasks are susceptible to learning dataset bias via surface cues; superficial cues that are associated with the label on a particular dataset. They investigated a recently proposed approach, called FLite \cite{le2020adversarial}. FLite adversarially filters dataset biases to mitigate the prevalent overestimation and overfitting of data in models. The authors demonstrate that FLite significantly reduces the measurable dataset biases, where models trained on the filtered datasets yielded better generalization to OOD tasks. However, their study stops short of extending its applicability to other language tasks, e.g., sentiment analysis, spam detection.

Clark \etal~\cite{clark2019don} argued that NLP models suffer from generalization and OOD issues due to biases in training datasets. Their study showed that the prior knowledge of this biases will enable training a model to be more robust to domain shift. The study demonstrated through experiments on several datasets with out-of-domain test sets huge robustness gains.

 Finally, Schiller \etal~\cite{schiller2021stance} introduced a stance detection benchmark, called  StD, to add and evaluate adversarial attack sets for NLP tasks. Their study demonstrated that the existence of biases inherited from multiple datasets by design leads to lack of robustness against adversarial examples. The authors stressed the need to focus on robustness and de-biasing strategies in multi-task learning approaches. However, the study did not offer recommendations on the applicability of this approach to other NLP tasks such as sentiment analysis, NLI, and question answering models.

\subsection{Spurious Correlation} 
  NLP models are generally prone to learning surface cues from training data which leads to the phenomena ``Spurious correlation''. Spurious correlations fool NLP models into making wrong predictions because models tend to rely on simple shortcuts rather than relying on the actual, typically complex, relationships in making such predictions. 
  
  The spurious correlation issue has been examined extensively in numerous research studies including the works in~\cite{moller2021informed, choe2020empirical, zhang2021causally, yaghoobzadeh2019increasing, wang2020identifying}. The way those research works tackle this issue varies greatly, depending on the linguistic task. For example, in \cite{yaghoobzadeh2019increasing}, Yaghoobzadeh  \etal presented a novel approach to design more robust NLP models and address the spurious correlation issue systematically. Their framework is based on {\em example forgetting}, where they find minority examples without any knowledge of the correlations present in the dataset. They tested their technique using three NLP taks (NLI, paraphrase identification, and fact verification) and showed consistent robustness gains. However, the study stops short of discussing other strategies to address the harm of spurious correlations such as data diversity and invariance.   

 Kaushik \etal~\cite{kaushik2019learning} studied whether NLP models pick up spurious patterns (e.g., if they are taking short-cuts instead of learning about the dataset when making predictions). They discovered that BERT and BiLSTM models  trained on original data fail to make correct predictions, while performing remarkably better when trained on combined datasets (counterfactually-revised counterparts). They have shown the results to generalize for multiple tasks; i.e., sentiment analysis and NLI. In their pursuit, the authors used humans in the loop to provide labels (predictions) and to intervene upon the data which may not be realistic given large datasets in the real-world. The study, however, did not offer any insights on how their models would perform on challenging adversarial datasets where spurious correlations do not necessarily hold. 
 
 Wang \etal~\cite{wang2020identifying} examined the effect of spurious correlation on the accuracy and robustness of text classifiers using a BERT model. They argue that NLP models are prone to learning surface cues during training, which may cause models to make incorrect predictions. They conducted studies using sentiment analysis NLP tasks. The study suggests feature engineering strategies to accomplish robustness, although it stops shorts of offering any insights on the transferibility and generalization of their approach to other architectures, e.g., CNN.

  


\subsection{Insights and Open Directions}
It is clear that rich benchmark datasets utilized in the literature for evaluating NLP models are diverse but inconsistent. Impressive results on benchmark datasets should not imply that a given NLP model will perform remarkably when deployed in the real-world. Research has shown that models are susceptible to solving the dataset rather than solving the underlying language understanding task \cite{jung2019earlier}. 

Overall, our exploration of the benchmark datasets space calls for work in various directions to fill various gaps. (1) While there is a significant initial work on the utilization of benchmark datasets that are free of bias in the broad NLP community, the datasets available so far are limited in many ways, and there is a need for developing techniques for identifying and removing biases in data to evaluate how NLP models would perform when deployed to the real-world. Namely, it would be interesting and worthwhile to extend the existing notions and benchmarks to techniques that address the existence of spurious correlation in datasets. (2) Using the same dataset for both training and testing might provide a false-sense of a model's robustness and accuracy. It would be interesting and worthwhile to test NLP models on various datasets by adopting the super GLUE, which is a successor of GLUE, a more challenging suite of datasets for various linguistic tasks. (3) There seems to be a gap and the need for a unified evaluation framework to enable comprehensive evaluations across various linguistic tasks in a fair and reproducible fashion. Namely, developing a standardized, unified evaluation benchmark dataset would be intriguing. In addition, modeling a strong set of baselines to be used as a test bed and trained on domain-specific data would also be interesting.

\section{Conclusion}\label{sec:conclusion}
This paper presents a survey on NLP robustness research in a consistent and systemic way.  We identified  various gaps in the literature with recommendations on future area of research directions following various elements in the NLP pipeline. As numerous real-world NLP projects have failed after deployment due to lack of robustness, exploring the robustness as a multi-dimensional concept that requires the development of new techniques is paramount. We note that newly developed techniques should address the spurious correlation challenges and achieve high out-of-distribution accuracy to ensure sufficient sensitivity to perturbations and ultimately lead to high precision in realistic text classification settings. Overall, it is our hope that this research work will serve as a fresh guide for the research community on technique, metric, and dataset to use, and motivate for additional interest and work in this space addressing the various gaps.

\bibliographystyle{IEEEtran}
\bibliography{ref.bib}

\end{document}